\documentclass[10pt,journal,compsoc,final]{IEEEtran}
\ifCLASSOPTIONcompsoc
  \usepackage[nocompress]{cite}
\else
  \usepackage{cite}
\fi
\ifCLASSINFOpdf
  \usepackage[pdftex]{graphicx}
  \graphicspath{{./Figures/}}
\else
\fi
\usepackage{amsmath}
\usepackage{algorithmic}
\usepackage{amssymb}
\usepackage{amsthm}
\usepackage{multirow}
\usepackage{bbm}
\usepackage{mathptmx}
\usepackage{bm}
\usepackage{url}

\DeclareMathOperator*{\argmin}{arg\,min}

\newcommand{\inv}[1]{\left({#1}\right)^{-1}}

\newcommand{\etal}{\textit{et al. }}
\usepackage{enumitem}
\usepackage{xcolor}

\newcommand{\mati}[1]{#1_{(:,i)}}
\begin{document}
%
\title{Relevance Subject Machine: A Novel Person Re-identification Framework}
%
%
%
%
\author{Igor Fedorov, \IEEEmembership{Student Member, IEEE,} Ritwik Giri, \IEEEmembership{Student Member, IEEE,} Bhaskar D. Rao, \IEEEmembership{Fellow, IEEE,} Truong Q. Nguyen, \IEEEmembership{Fellow, IEEE}%
\thanks{I. Fedorov, B.D. Rao and T.Q. Nguyen are with the Department of Electrical and Computer Engineering at the University of California, San-Diego. R. Giri was with the Department of Electrical and Computer Engineering at the University of California, San-Diego when the research for this work was conducted and is now with Starkey Hearing Technologies. I. Fedorov was partially supported by the San Diego Chapter of the ARCS Foundation, Inc.}}

\IEEEtitleabstractindextext{%
\begin{abstract}
We propose a novel method called the Relevance Subject Machine (RSM) to solve the person re-identification (re-id) problem. RSM falls under the category of Bayesian sparse recovery algorithms and uses the sparse representation of the input video under a pre-defined dictionary to identify the subject in the video. Our approach focuses on the multi-shot re-id problem, which is the prevalent problem in many video analytics applications. RSM captures the essence of the multi-shot re-id problem by constraining the support of the sparse codes for each input video frame to be the same. Our proposed approach is also robust enough to deal with time varying outliers and occlusions by introducing a sparse, non-stationary noise term in the model error. We provide a novel Variational Bayesian based inference procedure along with an intuitive interpretation of the proposed update rules. We evaluate our approach over several commonly used re-id datasets and show superior performance over current state-of-the-art algorithms. Specifically, for ILIDS-VID, a recent large scale re-id dataset, RSM shows significant improvement over all published approaches, achieving an $11.5 \%$ (absolute) improvement in rank $1$ accuracy over the closest competing algorithm considered.
\end{abstract}

}

\maketitle

\IEEEdisplaynontitleabstractindextext

%
\IEEEpeerreviewmaketitle

\IEEEraisesectionheading{\section{Introduction}\label{sec:introduction}}
\IEEEPARstart{C}{amera} networks have become increasingly commonplace in crowded public settings, such as airports and subways. These networks can be used as a tool for improving public safety by enabling the identification and tracking of target individuals \cite{de2009people}. We focus on the person re-identification (re-id) task, which refers to determining the identity of an individual in a video given a database of camera footage \cite{gong2014person}. The input video is often referred to as the probe and the database as the gallery \cite{karanam2015sparse}. The gallery may consist of videos recorded from the same camera as the probe video or from different cameras \cite{hamdoun2008person}. The re-id problem is challenging because subjects may be occluded in the probe and/or gallery, and there may be significant illumination and perspective variation between the gallery and probe.  

The defining characteristics of re-id algorithms are the representation of video frames (the feature space) and the method by which the probe is matched to the subjects in the gallery (the ranking system). There is already a significant body of work in defining highly robust and discriminative feature spaces for the re-id task \cite{wu2016deep,mclaughlin2016,lisanti2015person,gray2008viewpoint,niyogi2004locality,farenzena2010person,tuzel2006region,bkak2011multiple,liao2015person,
corvee2010person,zhao2013unsupervised,klaser2008spatio,hirzer2012person,cheng2011custom,bazzani2012multiple}.
In addition, much work has been done in studying ranking schemes for the case when the probe contains a single frame \cite{wright2009robust,wright2011sparsity,farenzena2010person,Cheng:BMVC11,lisanti2015person}. 
In many instances, researchers extend their ranking algorithms to the scenario where the probe contains \textit{multiple} frames \cite{karanam2015sparse,lisanti2015person} without exploiting a significant amount of information provided by the \textit{collection} of probe frames. 
The primary motivation for our work is to develop a ranking scheme specifically tailored for the multi-shot re-id problem. As such, we focus primarily on the design of the ranking strategy, assuming that the feature space has been pre-defined.

\begin{figure*}
\centering
\includegraphics[width=\textwidth]{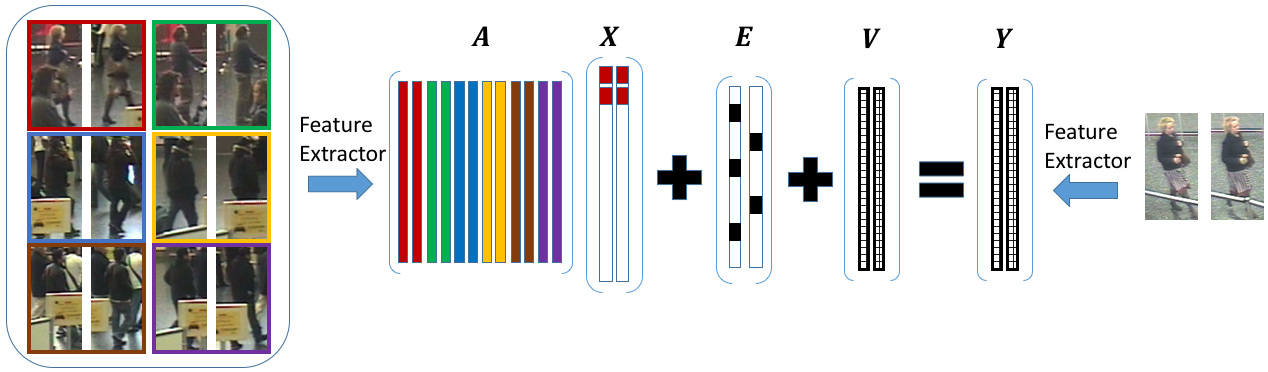}
\caption{Diagram of RSM signal model.}
\label{fig:model}
\end{figure*}

\subsection{Problem Formulation}
Let $\lbrace \bm{\mati{A}} \in \mathbb{R}^d \rbrace_{i=1}^N$ represent the set of gallery frames, in a pre-defined feature space, as shown in Fig. \ref{fig:model}. Note that each $\bm{\mati{A}}$ is associated with a subject labeled with index $\bm{s_i}$, $1 \leq \bm{s_i} \leq C $, where $C$ is the total number of subjects. Let $\lbrace \bm{\mati{Y}} \in \mathbb{R}^d \rbrace_{i=1}^L$ represent the set of frames for a given probe. The goal of re-id is to generate $\bm{\psi} \in \mathbb{R}^C$, where $\bm{\psi_r}$ is the rank $r$ estimate of the identity of the subject in the probe.

The re-id problem has two distinct variants: single or multi-shot \cite{farenzena2010person}. The number of probe frames $L$ is constrained to be $1$ for single-shot, whereas $L > 1$ for multi-shot. In both cases, the gallery $\bm{A}$ may contain one or more columns corresponding to each subject. 

Recently, sparsity based methods have become increasingly popular in computer vision contexts \cite{wright2009robust,elhamifar2009sparse}, including re-id \cite{lisanti2015person,karanam2015sparse,harandi2012sparse}. Specifically, the sparse representation-based classification (SRC) \cite{wright2009robust} framework has become a common-place building block in numerous applications. In the context of re-id, SRC can be seen as a way of addressing the single-shot problem. The idea is to first solve the following sparse recovery (SR) problem
\begin{align}\label{eq:l1}
\argmin_{\bm{x}} \underbrace{\Vert \bm{y} - \bm{A x} \Vert_2^2}_{\text{Modeling error}} +  \underbrace{\lambda\Vert \bm{x} \Vert_1}_{\text{Sparsity regularizer}},
\end{align}
where we use $\bm{y}$ to denote $\bm{Y_{(:,1)}}$ for brevity. 
We refer to $\bm{x}$ as the encoding of $\bm{y}$ under $\bm{A}$. The $\ell_1$ regularizer in \eqref{eq:l1} promotes sparse $\bm{x}$ \cite{eladbook,baraniuk2007compressive}, i.e. $\bm{x}$ that has many zeros. The hypothesis is that $\bm{y}$ lies in a subspace spanned by $\lbrace \bm{\mati{A}} \rbrace_{i: \bm{s_i} = c^*}$, where $c^*$ is the identity of the subject in $\bm{y}$. Under this hypothesis, the support of $\bm{x}$ can be treated as a good indicator of the identity of $\bm{y}$. To generate $\bm{\psi}$, the residuals
\begin{align}\label{eq:src classifier}
\bm{r_c^{SRC}} = {\Vert \bm{y} - \bm{A} \phi_c(\bm{x}) \Vert_2}
\end{align}
are first computed for all subjects $c$, where $\phi_c(\cdot)$ is defined such that, if $\bm{z} = \phi_c(\bm{x})$, then
\begin{equation}
\bm{z_j} =  \begin{cases} 
      \bm{x_j} & \text{if } \bm{s_j} = c \\
      0 & \text{else.} 
   \end{cases}
\end{equation}
The elements of $\bm{r^{SRC}}$ are then sorted in ascending order and $\bm{\psi}$ is set to $\bm{\rho}$, where $\bm{r_{\rho_r}^{SRC}}$ is the $r$'th smallest element of $\bm{r^{SRC}}$.

While effective, the formulation in \eqref{eq:l1} is sub-optimal for the person re-id problem in many respects. One significant issue is that \eqref{eq:l1} ignores the inherent structure in $\bm{A}$ when solving the SR problem. As Fig. \ref{fig:model} shows, $\bm{A}$ has a \textit{block} structure, where each block $g_c = \lbrace 1 \leq i \leq N: \bm{s_i} = c \rbrace$ corresponds to a specific subject in the gallery \cite{karanam2015sparse}. Given that $\bm{A}$ has a block structure, $\eqref{eq:l1}$ can be modified to solve a \textit{block sparse recovery} problem \cite{eldar2010block,zhang2013extension} by replacing the $\ell_1$ regularizer with a block-sparsity regularizer.

Another issue with \eqref{eq:l1} is that it is sensitive to large outliers in the modeling error, which may occur because of occlusions, illumination changes, viewpoint variations, etc. To make the formulation more robust, the modeling error is often modified to include a dense (Gaussian) noise term \textit{and} a sparse error term \cite{wright2009robust}, represented by $\bm{V}$ and $\bm{E}$ in Fig. \ref{fig:model}, respectively. 


Most importantly, although \eqref{eq:l1} is a reasonable approach to the single-shot re-id problem, extending it to the multi-shot case is not straightforward. There have been several attempts at building upon the SRC framework for the multi-shot problem \cite{lisanti2015person,karanam2015sparse}.
The issue with these methods is that they neglect to exploit the fundamental advantage offered by the multi-shot re-id problem:  The identity of the probe subject is constant for all probe frames $\lbrace\bm{\mati{Y}}\rbrace_{i=1}^L$, meaning that $\lbrace \bm{\mati{X}} \rbrace_{i=1}^L$ should exhibit \textit{joint} sparsity, i.e. the sparsity pattern of $\bm{\mati{X}}$ should be the same for all $i$, as shown in Fig. \ref{fig:model}. Incorporating joint sparsity into the SR algorithm is known to produce superior results \cite{wipf2007empirical}. The lack of a re-id algorithm capable of exploiting joint sparsity motivates the present work.

\subsection{Contribution}
We propose a re-id framework which incorporates many of the advantages of previous SR based approaches, while also leveraging the information inherent in the multi-shot problem to a fuller extent.

\begin{itemize}
\item We employ a Bayesian approach to the SR problem, based on the Relevance Vector Machine (RVM) \cite{tipping2001sparse}. RVM has been shown to achieve much higher SR rates than many deterministic approaches \cite{giri2015type} and has the potential to outperform algorithms based on the formulation in \eqref{eq:l1} in the re-id task.  We build upon the signal model we introduced in \cite{fedorov2016robust} and refer to our approach as the Relevance Subject Machine (RSM).

\item Unlike \cite{karanam2015sparse,lisanti2015person}, RSM allows us to use all of the frames in the probe \textit{within} the SR process and model the joint sparsity of $\lbrace \bm{\mati{X}} \rbrace_{i=1}^L$. 
In addition, RSM allows us to model sparse, outlier noise prevalent in the re-id problem.


\item We introduce a novel iterative candidate subject ranking strategy which leverages the superior SR performance of the Bayesian approach without making inference computationally intractable.

\item We show that the proposed approach achieves state-of-the-art performance on multi-shot re-id datasets.
\end{itemize}

\section{Related Work}
\subsection{State-of-the-Art Sparse Re-Identification}
\label{sec:State of the art}
SR based re-id algorithms are generally characterized by their representation of the video data, SR algorithm, and ranking scheme. A thorough review of popular feature spaces for the re-id problem can be found in \cite{karanam2016comprehensive}. Usually, a tracking algorithm is used to isolate a single subject in each frame. The remaining design choice is then to select the representation of the subject. Gray and Tao \cite{gray2008viewpoint} split each frame into $6$ horizontal strips, convolve each strip with $13$ Schmid and $6$ Gabor filters, and form a feature vector by quantizing the filter responses into 16 bin histograms. The feature vector is then appended with $16$ bin histograms of the RGB, HSV, and YCbCr representations of each strip, resulting in a $2592$-dimensional representation. Lisanti \etal \cite{lisanti2015person} split each frame into overlapping horizontal stripes and extract RGB and Hue-Saturation histograms for each stripe, where the contribution of each pixel is weighted by an Epanechnikov kernel centered on the image. The feature vector is then appended with a Histogram of Oriented Gradient (HOG) descriptor computed over the entire frame, generating a $2960$-dimensional representation referred to as the weighted histogram of overlapping stripes (WHOS). Liao \etal \cite{liao2015person} describe the local maximal occurrence (LOMO) representation, where a $26960$-dimensional representation is computed using a set of local horizontal features. 
Other feature spaces based on salience \cite{zhao2013unsupervised}, symmetry driven accumulation of local features (SDALF) \cite{farenzena2010person}, and local binary patterns (LBP) \cite{hirzer2012relaxed} have been explored. In most cases, the number of features is \textit{greater} than the number of gallery frames. This motivates the approach in the present work, whose computational complexity is a function of the \textit{minimum} of $d$ and $N$.

In order to aid in classification, many approaches learn a projection (referred to as a metric) such that projected points from the same class are close to each other and far away from projected points from other classes \cite{weinberger2005distance}. Karanam \etal \cite{karanam2015sparse} use local Fisher discriminant analysis \cite{sugiyama2006local} to learn an embedding given training data for both the gallery and probe views. Liao \etal \cite{liao2015person} present an efficient metric learning algorithm, called Cross-view Quadratic Discriminant Analysis (XQDA), aimed at learning projections when features are extracted from two separate views.

SR algorithms for re-id usually include solving a problem of the form given in \eqref{eq:l1}. Khedher \etal \cite{khedher2013multi} use \eqref{eq:l1} to extract sparse codes for each frame of a probe. Karanam \etal \cite{karanam2015sparse} propose the sparse re-id (SRID) framework where a block-sparse, outlier robust SR approach of the form
\begin{equation}\label{eq:block l1}
\begin{aligned}
& \underset{\bm{\mati{X}},\bm{\mati{E}}}{\text{argmin}}
& & \Vert \bm{\mati{E}} \Vert_1  + \sum_{c=1}^C \Vert \bm{X_{(g_c,i)}} \Vert_2 \\
& \text{subject to}
& & \bm{\mati{Y}} = \bm{A \mati{X}}+\bm{\mati{E}}
\end{aligned}
\end{equation}
is used for each probe frame $i$, where $\bm{X_{(g_c,i)}}$ is shorthand for $\begin{bmatrix}
\bm{X_{(j_1,i)}} & \cdots & \bm{X_{(j_{\vert g_c \vert} , i)}}
\end{bmatrix}^T$, $j_k \in g_c$. 
Lisanti \etal \cite{lisanti2015person} propose an iterative sparse re-weighting (ISR) approach for the re-id problem where, at each iteration $r$, a problem of the form
\begin{align}\label{eq:rl1}
\argmin_{\bm{\mati{X}}} \Vert \bm{\mati{Y}} - \bm{A \mati{X}} \Vert_2^2 + \lambda \Vert \bm{W^r} \bm{\mati{X}} \Vert_1
\end{align}
is solved for each probe frame $i$, where $\bm{W^r}$ is a diagonal matrix whose $\left(j,j\right)$'th entry is given by $\inv{\bm{X_{(j,i)}^{r-1}}+\epsilon}$, $\epsilon$ is a small positive constant, and $\bm{\mati{X^r}}$ denotes the estimate of $\bm{\mati{X}}$ at the $r$'th iteration. The intuition for performing reweighting is to make the sparse-recovery algorithm robust against large non-zero values of $\bm{\mati{X}}$ \cite{candes2008enhancing}, which heavily influence the $\ell_1$ norm and can prevent formulations such as \eqref{eq:l1} from finding smaller valued coefficients that can be important for ranking. RSM incorporates the advantages of \eqref{eq:block l1} \textit{and} \eqref{eq:rl1}. Moreover, unlike previous SR based re-id approaches based on \eqref{eq:l1}, \eqref{eq:block l1}, and \eqref{eq:rl1} \cite{karanam2015sparse,lisanti2015person,khedher2013multi,harandi2012sparse}, which solve the SR problem for each probe frame \textit{independently}, we perform SR on the entire probe video under a joint sparsity assumption. %

Ranking schemes used by SR based re-id algorithms usually take the form of computing a residual for each subject $c$, as in \cite{karanam2015sparse} where
\begin{align}\label{eq:classifier karanam}
\bm{r_{c}^{SRID}} = \sum_{i=1}^L \Vert \bm{\mati{Y}} - \bm{\mati{E}} - \bm{A} \phi_c(\bm{\mati{X}}) \Vert_2,
\end{align}
and sorting the result to produce $\bm{\psi}$.
One variant of this approach is given in \cite{lisanti2015person}, where the residual computation is modified to
\begin{align}\label{eq:residual isr}
\bm{r_c^{ISR}} = \argmin_i \Vert \bm{\mati{Y}} - \bm{A} \phi_c(\bm{\mati{X}}) \Vert_2
\end{align}
and the ranking is generated iteratively. At each iteration $r$, a SR problem of the form \eqref{eq:rl1} is solved for each probe frame $i$, the residuals in \eqref{eq:residual isr} computed, and $\bm{\psi_r}$ set to $\argmin_c \bm{r_c^{ISR}}$. Before proceeding to the next iteration, columns corresponding to subject $\bm{\psi_r}$ are removed from $\bm{A}$ in order to prevent subjects which have already been ranked from influencing the ranking of the remaining subjects. We employ the residual definition in \eqref{eq:classifier karanam} because it encodes the reconstruction error for all $L$ probe frames, which is more aligned with our goal of \textit{jointly} identifying $\bm{Y}$. In addition, we build upon the iterative ranking strategy in \cite{lisanti2015person} in our RSM framework.


\subsection{Bayesian Sparse Recovery}
\label{sec:BSR}

Recently, there has been much interest in Bayesian approaches for the SR problem. Tipping's seminal work \cite{tipping2001sparse} on the RVM offered a novel hierarchical Bayesian formulation for sparse regression and classification problems. Wipf and Rao \cite{wipf2004sparse} then adopted the RVM for the SR problem, showing that, in the noiseless setting, the local minima of the objective function being minimized are sparse and the global minimum is the \textit{maximally sparse} encoding of $\bm{y}$ under $\bm{A}$. Moreover, RVM has been shown to empirically outperform competing SR algorithms \cite{giri2015type}.

In the following, let $\bm{y}$ and $\bm{x}$ be vector random variables representing a single probe frame and its corresponding sparse code, respectively, under $\bm{A}$. The RVM signal model is given by
\begin{align}\label{eq:signal model}
\bm{y} = \bm{A} \bm{x} + \bm{v}
\end{align}
where $\bm{v} \sim {N}(0,\lambda \bm{\mathsf{I}})$ represents signal noise. The sparse code $\bm{x}$ is assigned a separable prior of the form $p(\bm{x}) = \prod_{j=1}^N p(\bm{x_j})$
where $p(\bm{x_j})$ has a Gaussian Scale Mixture (GSM) \cite{andrews1974scale} representation given by
\begin{align}\label{eq:GSM}
p(\bm{x_j} | \bm{\gamma_j}) =  {N}(\bm{x_j};0,\bm{\gamma_j}).
\end{align}
The scale mixture representation is capable of representing most heavy-tailed densities \cite{palmer2006variational,palmer2005variational,lange1993normal,dempster1980iteratively,dempster1977maximum}, which are known to be suitable for promoting sparsity \cite{tipping2001sparse,wipf2007empirical}. The heavy-tailed nature of scale mixture priors allows us to model both the sparsity and large non-zero entries of $\bm{x}$. Note that the selection of $p(\bm{\gamma})$ represents a choice of prior on $\bm{x}$ \cite{wipf2004sparse}. The RVM framework seeks the maximizer $\bm{\gamma^{*}}$ of the evidence $p(\bm{y} | \bm{\gamma})$ and then estimates $\bm{x}$ using the mode of $p(\bm{x} | \bm{y} , \bm{\gamma^{*}})$.

One of the advantages of the RVM is that it is straightforward to encode prior knowledge about the recovery problem. For instance, if $\bm{A}$ is known to be decomposable into disjoint blocks, the conditional density of $\bm{x}$ given $\bm{\gamma}$ is amended to \cite{zhang2011sparse}
\begin{align}\label{eq:block GSM}
p(\bm{x} | \bm{\gamma}) = \prod_{c=1}^C \prod_{j \in g_c}  {N}(\bm{x_j};0,\bm{\gamma_c}) 
\end{align}
and the inference procedure remains the same. Note that, whereas $\bm{\gamma} \in \mathbb{R}^N$ in \eqref{eq:GSM}, $\bm{\gamma} \in \mathbb{R}^C$ in \eqref{eq:block GSM}. To additionally encode joint sparsity among the columns of $\bm{X}$, we can define a \textit{matrix} conditional density of the form \cite{wipf2007empirical}
\begin{align}\label{eq:mmv sbl}
\begin{split}
p(\bm{X} | \bm{\gamma}) &= \prod_{c=1}^C \prod_{j \in g_c} {N}(\bm{X_{(j,:)}} ; 0 , \bm{\gamma_c \mathsf{I}}).
\end{split}
\end{align}
Joint sparsity is enforced in \eqref{eq:mmv sbl} by the fact that $\bm{\gamma_c}$ is shared for all $j \in g_c$ among all $\lbrace \bm{X_{(:,i)}} \rbrace_{i=1}^L$. Finally, incorporating robustness to sparse noise is achieved by modifying \eqref{eq:signal model} to \cite{li2013robust}
\begin{align}
\bm{y} = \bm{A} \bm{x} + \bm{v} + \bm{e}
\end{align}
where $\bm{e}$ is itself assigned a separable GSM prior to promote sparsity.

The combination of block sparsity, robustness to sparse noise, and joint sparsity over multiple observations was first studied by us in \cite{fedorov2016robust}. The signal model considered was
\begin{align}\label{eq:robust sbl}
\bm{Y} =  \bm{A}\bm{X} + \bm{V} + \bm{E},
\end{align}
where $p(\bm{X}|\bm{\gamma})$ is given by \eqref{eq:mmv sbl} and $p(\bm{E} | \bm{D})$ is given by


\begin{equation}
p (\bm{E} | \bm{D}) = \prod_{j=1}^d \prod_{i=1}^L {N}(\bm{E_{(j,i)}} ; 0, \bm{D_{(j,i)}}).
\end{equation}
Although the joint block sparsity constraint is a valid one for the encoding matrix $\bm{X}$, it does not hold for the outlier matrix $\bm{E}$ since the outliers can be time varying. 
Because the Inverse Gamma (${IG}$) distribution is the conjugate prior for the variance of a Gaussian likelihood,  we specify $p(\bm{\gamma_c}) = {IG} (\alpha_{\gamma}, \beta_{\gamma})$ and $p(\bm{D_{(j,i)}}) = {IG} (\alpha_{\delta}, \beta_{\delta})$.

To the best of our knowledge, we are the first to apply the RVM framework to person re-id. One reason why previous researchers may have been reluctant to apply the RVM to re-id may be that inference was computationally intractable. To combat this, 
we propose a ranking strategy based on \cite{lisanti2015person} which can be readily used within our RSM framework without adding a significant amount of additional computation. Finally, we apply metric learning to reduce the dimensionality of the problem, where appropriate.

\section{Approach}
In this section, we describe the inference and ranking strategies in detail. We begin by introducing Variational Bayesian (VB) inference in Section \ref{sec:VB background}. We show the VB inference procedure for RSM in Section \ref{sec:VB for RSM} and the ranking strategy in Section \ref{sec:ranking}.

\subsection{Variational Bayesian Inference Background}
\label{sec:VB background}
Let $\bm{z}$ and $\bm{y}$ denote the set of all latent and observed variables, respectively, in a given model. To estimate the set of desired parameters, which are included in $\bm{z}$,  the goal is to find the posterior density $p(\bm{z} | \bm{y})$. The primary hindrance to computing this posterior lies in the integration required to find the normalizing factor $ p(\mathbf{y}) = \int p(\mathbf{y}|\mathbf{z})  p (\mathbf{z}) d \mathbf{z}$, which makes computing $p(\bm{z} | \bm{y})$  intractable for most densities of interest. In VB \cite{mackay1995ensemble,jordan1999introduction,bishop2006pattern}, this problem can be bypassed by approximating $p(\bm{z} | \bm{y})$ by a factorized distribution $q(\bm{z}) = \prod_{k=1}^K q_k(\bm{z_k})$ 
 where $\bm{z_k}$ denotes a subset of $\bm{z}$. Consider the log marginal probability of $\bm{y}$
\begin{equation}
\log p(\bm{y}) = {L}(q(\bm{z})) + KL \left(q(\bm{z}) \; || \; p(\bm{z} | \bm{y})\right),
\label{eq:marginal}
\end{equation}
where the first term is defined as
\begin{align}
{L}(q) = \int q(\bm{z}|\bm{y}) \log \frac{p(\bm{y}, \bm{z})}{q(\bm{z}|\bm{y})} d \bm{z}
\end{align}
and can be interpreted as the lower bound on the log marginal probability, while the second term is the Kullback-Leibler (KL) divergence between $q$ and the true posterior. The aim is to maximize the lower bound with respect to $q$, which is equivalent to minimizing the KL divergence term.  

Let us now select one subset $\bm{z_k}$ and maximize ${L}(q)$ with respect to $q_k$. It can be shown \cite{mackay1995ensemble,jordan1999introduction,bishop2006pattern} that this optimization leads to the following update of $q_k$:
\begin{equation}\label{eq:VB definition}
\log q_k = E_{k' \neq k} \left[ \log p(\bm{y},\bm{z}) \right]
\end{equation}
where $E_{k'\neq k} [\cdots]$ denotes an expectation with respect to $q$ over all variables $\bm{z_{k'}}$, $k'\neq k$. This procedure is then iteratively repeated for all subsets to convergence \cite{boyd2004convex}.

\subsection{Variational Bayesian Inference for the RSM}
\label{sec:VB for RSM}
We now turn to the task of performing VB inference for the RSM signal model. The signal model employed by the RSM framework is shown in \eqref{eq:robust sbl} and visualized in Fig. \ref{fig:model}. In the following, let $\bm{z} = \left(\lbrace \bm{\mati{X}} \rbrace_{i=1}^L,\lbrace \bm{\mati{E}} \rbrace_{i=1}^L, \lbrace \bm{\gamma_c} \rbrace_{c=1}^C,\lbrace \bm{D_{(j,i)}} \rbrace_{j=1,i=1}^{j=d,i=L} \right)$ be the set of latent variables.

\subsubsection{Computation of $q(\bm{X_{(:,i)}})$} 
\label{sec: computation of qx}
Using \eqref{eq:VB definition}, we find that $q(\bm{\mati{X}}) = {N}(\bm{\mati{X}}; \bm{\mati{\mu}^x},\bm{\Sigma^x})$, where
\begin{align}
\bm{\Sigma^x} &= \inv{ \inv{\lambda} {\bm{A}^T \bm{A}} + \inv{\bm{\Gamma}}} \label{update_sigma_x}\\
\bm{\mati{\mu}^x} &= \inv{\lambda}{\bm{\Sigma^x} \bm{A}^T} \left(\bm{\mati{Y}}-\langle \bm{\mati{E}} \rangle \right) \label{update_mu_x}
\end{align}
and $\bm{\Gamma}$ is a diagonal matrix with the $(j,j)$'th entry given by $\left\langle \frac{1}{\bm{\gamma_c}} \right\rangle$, for $j \in g_c$. We use $\langle \bm{z_k} \rangle$ to denote the expectation of the random variable $\bm{z_k}$ with respect to $q_{k}$. Note that $\lbrace q(\bm{\mati{X}}) \rbrace_{i=1}^L$ share the same covariance matrix $\bm{\Sigma^x}$. 

\subsubsection{Computation of $q(\bm{\mati{E}})$}
Using \eqref{eq:VB definition}, we find that $q(\bm{\mati{E}}) = {N}(\bm{\mati{E}};\bm{\mati{\mu}^e},\bm{\Sigma_i^e})$, where
\begin{align}
\bm{\Sigma_i^e} &= \inv{\inv{\lambda}{\bm{\mathsf{I}}}+\bm{\Delta_i}} \label{eq:Sigma E}\\
\bm{\mati{\mu}^e} &= \inv{\lambda}\bm{\Sigma_i^e}\left({\bm{\mati{Y}}}-\bm{A}\langle \bm{\mati{X}} \rangle\right)\label{eq:mu E}
\end{align}
and $\bm{\Delta_i}$ is a diagonal matrix with the $(j,j)$'th entry equal to $\left\langle \frac{1}{\bm{\bm{D}_{(j,i)}}} \right\rangle$.

\subsubsection{Computation of $q(\bm{\gamma_c})$ and $q(\bm{D_{(j,i)}})$}
\label{sec: computation of qd}
Using \eqref{eq:VB definition}, we find that $q(\bm{\gamma_c}) = {IG}\left(\bm{\gamma_c}; \frac{L \vert g_c\vert}{2}+\alpha_{\gamma},\frac{\langle \Vert \bm{X_{(g_c,:)}} \Vert_F^2 \rangle}{2} + \beta_{\gamma}\right)$ and $q(\bm{D_{(j,i)}}) = {IG}\left(\bm{D_{(j,i)}}; \frac{1}{2}+\alpha_{\delta},\frac{\langle \bm{E_{(j,i)}}^2 \rangle}{2} + \beta_{\delta}\right)$. 

\subsection{Summary}
\label{sec:ranking}
It is evident from Sections \ref{sec: computation of qx}-\ref{sec: computation of qd} that all of the $q_k$  depend on the moments of one or more latent variables. We will use an iterative procedure where we initialize the moments and, at each iteration, update the estimate of $q_k$ based on the current estimates of $\lbrace q_{k'} \rbrace_{k' \neq k}$, repeating the process for all $k$ until convergence. For clarity, we summarize the required moments for our iterative procedure:
\begin{align}
\left \langle \frac{1}{{\bm{\gamma_c}}} \right\rangle &= \frac{\frac{L \vert g_c \vert}{2}+\alpha_{\gamma}}{\frac{\langle \Vert \bm{X_{(g_c,:)}} \Vert_F^2 \rangle}{2} + \beta_{\gamma}}\label{eq:gamma_update}\\
\left \langle \frac{1}{{\bm{D_{(j,i)}}}} \right\rangle&=\frac{\frac{1}{2}+\alpha_{\delta}}{\frac{\langle \bm{E_{(j,i)}}^2 \rangle}{2} + \beta_{\delta}} \\ 
 \langle \bm{\mati{E}} \rangle &= \bm{\mati{\mu}^e} \; \; \; , \; \; \langle \bm{\mati{X}} \rangle = \bm{\mati{\mu^x}}\\
 \langle \bm{E_{(j,i)}}^2 \rangle &= \left(\bm{\mu_{(j,i)}^e}\right)^2+\bm{\left(\Sigma_i^e \right)_{(j,j)}} \\
 \langle \Vert \bm{X_{(g_c,:)}} \Vert_F^2 \rangle &= \Vert \bm{M_{(g_c,:)}} \Vert_F^2 + L \times \text{trace}\left(\bm{\Sigma^x_{(g_c,:)}} \right)\label{eq:Xblock update}
\end{align}
where $\bm{M} = \begin{bmatrix}
\bm{\mu^x_{(:,1)}} & \cdots & \bm{\mu^x_{(:,L)}}
\end{bmatrix}$. 
Although it is possible to estimate $\lambda$ within the Bayesian framework, we choose to select it using cross validation.

The remaining design element of RSM is the choice of ranking strategy. Given a probe video, we begin by performing the VB inference procedure until convergence. We then use $\bm{\mu^x}$ and $\bm{\mu^e}$ as point estimates of $\bm{X}$ and $\bm{E}$, respectively, compute residuals for each subject in the gallery using \eqref{eq:classifier karanam}, and set $\bm{\psi_1}$ to the subject which achieves the minimum residual. Next, we remove columns of $\bm{A}$ corresponding to $\bm{\psi_1}$ and repeat the process. The motivation for this strategy comes from \cite{lisanti2015person}, where Lisanti \etal argue that the coefficients in $\bm{X}$ are dominated by a few subjects in the gallery. By removing the contribution of subject $\bm{\psi_r}$ from $\bm{A}$ and re-estimating $\bm{X}$, we hope that $\bm{\psi_{r+1}}$ will be more accurate because $\bm{X}$ will contain more information about subjects which \textit{have not} been ranked yet.

Since re-estimating $\bm{X}$ for $C$ subjects can be computationally burdensome, we introduce a slight optimization shortcut. While performing inference for rank $r$, we save the state of the algorithm at $\zeta T$ iterations, where $T$ is the total number of iterations and $0 < \zeta \leq 1$. Then, we initialize the inference computations for rank ${r+1}$ with the saved state and set $T$ to $\tau T$, where $0 < \tau \leq 1$. The number of iterations needed to compute $\bm{\psi_r}$ decays geometrically with increasing $r$, providing a considerable decrease in computational cost. At the same time, performance loss is mitigated by the "warm" initialization strategy. The complete algorithm is summarized in Alg. \ref{alg: complete}.

\begin{figure}
\begin{algorithmic}[1]
\REQUIRE $\bm{Y},\bm{A},\lbrace g_c \rbrace_{c=1}^C,\lambda,T,\zeta,\tau$
\STATE Initialize $\Omega = \lbrace 1,\cdots,C \rbrace, \bm{\gamma_c} = 1, \bm{\mu_{(j,i)}^x} = 1, \bm{D_{(j,i)}} = 1 \; \; \forall c,j,i$ 
\FOR{$r = 1$ to $C$}
\FOR{$t = 1$ to $T$}
\STATE $\bm{\mu^0} \leftarrow \bm{\mu^x}$
\STATE Update parameters of $q(\bm{\mati{X}})$, $q(\bm{\gamma_c})$, $q(\bm{\mati{E}})$, $q(\bm{D_{(j,i)}})$ $\forall c,j,i$
\STATE Update moments in \eqref{eq:gamma_update}-\eqref{eq:Xblock update}
\IF{$t == \zeta T$}
\STATE Save moments
\ENDIF
\ENDFOR
\STATE Compute $\bm{r_c^{SRID}}$ using \eqref{eq:classifier karanam} for all $c \in \Omega$
\STATE Find $c^* = \argmin_c \bm{r_c^{SRID}}$, 
\STATE Set $\bm{\psi_r} \leftarrow c^*$ and $\Omega \leftarrow \Omega \setminus c^*$
\STATE Remove columns of $\bm{A}$ corresponding to $c^*$
\STATE $T \leftarrow \tau T$
\STATE Re-initialize moments using saved values, removing any elements corresponding to $c^*$
\ENDFOR
\STATE 
\RETURN 
\end{algorithmic}
\caption{RSM algorithm}
\label{alg: complete}
\end{figure}

\section{Interpretation of Updates}
\label{sec:intepretation}
Consider the point estimate $\bm{\mati{\mu}^x}$ of $\bm{\mati{X}}$, formed after the proposed VB procedure converges.
Substituting \eqref{update_sigma_x} in \eqref{update_mu_x}, we get
\begin{align}
\bm{\mati{\mu}^x} = \inv{\bm{A}^T \bm{A} + \lambda \bm{\Gamma}} \bm{A}^T (\bm{\mati{Y}} - \bm{\mati{\mu}^e}).
\label{eq:estimate} 
\end{align}
We see that \eqref{eq:estimate} is actually the solution of a weighted ridge regression problem, i.e. least squares fitting with weighted $\ell_2$ norm regularization over the coefficient vector:
\begin{align}
\begin{split}
\bm{\mati{\mu}^x} =& \arg \min_{\bm{\mati{X}}} \Vert \bm{\mati{Y}} - \bm{\mati{\mu^e}}- \bm{A} \bm{\mati{X}} \Vert_2^2+ \lambda \Vert \bm{\Gamma^{\frac{1}{2}}} \bm{\mati{X}} \Vert_2^2 
\end{split}
\label{eq:alt}
\end{align}
where $\bm{\Gamma^{\frac{1}{2}}}$ is a diagonal matrix whose $(j,j)$'th entry is $\left( \langle \frac{1}{\bm{\gamma_c}} \rangle \right)^{\frac{1}{2}}$, for $j \in g_c$. The least squares fitting is done after removing the outlier estimate $\bm{\mati{\mu^e}}$ from the observation $\bm{\mati{Y}}$, which makes our approach very robust.

Hence, we can relate the RSM to ISR \cite{lisanti2015person}, which uses a reweighted $\ell_1$ formulation, shown in \eqref{eq:rl1}, and reports state-of-the-art results for a number of re-id datasets. Our inference procedure can be interpreted as a robust reweighted $\ell_2$ norm minimization algorithm, where the weights are computed in the hyperparameter space, known as Type II inference. In the SR literature, researchers have shown the superiority of Type II methods over methods which perform inference in the $\bm{X}$ space, which are termed Type I and include ISR and SRID \cite{giri2015type}. This observation motivated us to use a Type II based inference procedure for the re-id problem. As shown in Section \ref{sec:results}, our Type II approach shows significant improvement over ISR and SRID, as well as other re-id algorithms not based on SR.

\begin{table*}
\tiny
\begin{center}
\begin{tabular}{|cc|cccc|cccc|}
\hline
& & \multicolumn{4}{|c|}{ILIDS-VID} & \multicolumn{4}{|c|}{PRID 2011}\\
Classifier & Feature space & Rank 1 & Rank 5 & Rank 10 & Rank 20 & Rank 1 & Rank 5 & Rank 10 & Rank 20\\ \hline
RSM & LOMO + XQDA \cite{liao2015person} & \bm{$96.6$} & \bm{$99.6$} & \bm{$99.9$} & \bm{$99.9$ } & \bm{$99.4$} & \bm{$99.9$} & \bm{$100$} & \bm{$100$}\\
ISR \cite{lisanti2015person} & LOMO+ XQDA \cite{liao2015person} & $85.1$  & $95.2$ & $97.2$ & $98.5$ & {$99.1$} & {$99.8$} & {$99.9$} & \bm{$100$}\\
SRID \cite{karanam2015sparse} & Gray and Tao \cite{gray2008viewpoint} + LFDA \cite{niyogi2004locality} & $24.9$ & $44.5$ & $55.6$ & $66.2$ & $35.1$ & $59.4$ & $69.8$ & $79.7$ \\ \hline
TAPR\cite{gao2016temporally} & LOMO \cite{gao2016temporally} & $55$ & $87.5$ & $93.8$ & $97.2$ & $68.6$ & $94.6$ & $97.4$ & $98.9$ \\
Softmax & Deep RNN \cite{mclaughlin2016} & $58$ & $84$ & $91$ & $96$ & $70$ & $90$ & $95$ & $97$\\
NN & Deep CNN \cite{wu2016deep} & $42.6$ & $70.2$ & $86.4$ & $92.3$ & $49.8$ & $77.4$ & $90.7$ & $94.6$\\
TDL \cite{you2016top}&  HOG3D \cite{klaser2008spatio} & $56.3$ & $87.6$ & $95.6$ & $98.3$ & $56.7$ & $80$ & $87.6$ & $93.6$\\
NN & Salience \cite{zhao2013unsupervised} & $10.2$ & $24.8$ & $35.5$ & $52.9$ & $25.8$ & $43.6$ & $52.6$ & $62$ \\
RPRF \cite{li2015multi} & Gray and Tao \cite{gray2008viewpoint} & $14.5$ & $29.8$ & $40.7$ & $58.1$ & $19.3$ & $38.4$ & $51.6$ & $68.1$ \\
DVR \cite{wang2014person} & HOD3D \cite{klaser2008spatio} & $23.3$ & $42.4$ & $55.3$ & $68.4$ & $28.9$ & $55.3$ & $65.5$ & $82.8$\\
RankSVM \cite{chapelle2010efficient} & Color \& LBP \cite{hirzer2012relaxed} & $23.2$ & $44.2$ & $54.1$ & $68.8$ & $34.3$ & $56$ & $65.5$ & $77.3$ \\
NN & SDALF \cite{farenzena2010person} & $6.3$ & $18.8$ & $27.1$ & $37.3$ & $5.2$ & $20.7$ & $32$ & $47.9$ \\
\hline
\end{tabular}
\end{center}
\caption{CMC results for ILIDS-VID and PRID2011 datasets.}
\label{table:ilidsvid}
\end{table*}

\begin{table}
\tiny
\begin{center}
\begin{tabular}{|cc|cccc|}
\hline
Classifier & Feature space & Rank 1 & Rank 5 & Rank 10 & Rank 20 \\ \hline
RSM & WHOS \cite{lisanti2015person} & \bm{$95.8$} & \bm{$98.6$} & \bm{$100$} & \bm{$100$}   \\
ISR \cite{lisanti2015person} & WHOS \cite{lisanti2015person} & $90.1$  & $97.6$ & $99$ & $99.7$ \\
\hline
Bayes & CPS \cite{cheng2011custom} & $17.5$ & $47.5$ & $68$ & $86$\\
NN & SDALF \cite{farenzena2010person} & $8.3$ & $37.5$ & $58$ & $77.5$ \\
Bayes & AHPE \cite{bazzani2012multiple} & $7.5$ & $32$ & $65$ & $72$ \\ \hline
\end{tabular}
\end{center}
\caption{CMC results for CAVIAR4ReID dataset.}
\label{table:caviar}
\end{table}

Another key advantage of RSM lies in its ability to capture the essence of multi-shot data. We can see from \eqref{eq:alt} that the weight matrix $\bm{\Gamma^{\frac{1}{2}}}$ is shared among all $L$ probe frames. The weighting scheme combines information from all $L$ frames and takes advantage of the key fact that they belong to the same subject. Similarly, the moment computation for $\frac{1}{\bm{\gamma_c}}$ in \eqref{eq:gamma_update} combines information from all $L$ frames.

\section{Results}
\label{sec:results}
In this section, we report experimental results on various re-id datasets. To highlight the capabilities of RSM, we focus on multi-shot re-id datasets, including ILIDS-VID \cite{wang2016person}, PRID 2011 \cite{hirzer2011person}, and CAVIAR4ReID \cite{Cheng:BMVC11}. In addition, we also present results for a single-shot re-id task on the VIPeR dataset \cite{gray2007evaluating} in order to show that RSM is competitive with other single-shot re-id algorithms. We report results in terms of the cumulative matching characteristic (CMC). In addition, we use $\alpha_\gamma = \alpha_\delta = \beta_\gamma = \beta_\delta = 0$.

\subsection{Performance on ILIDS-VID}
\label{sec:ILIDSVID}
The ILIDS-VID dataset contains $300$ subjects with sequences from two camera views, with $23-192$ frames per subject per view. The dataset exhibits significant lighting, viewpoint, and occlusion variations and is ideal to showcase the capabilities of RSM. We elected to use the LOMO \cite{liao2015person} feature space, which has been shown to produce state-of-the-art results on a number of re-id datasets. Because of the large amount of data available in ILIDS-VID, we also learned the XQDA \cite{liao2015person} metric, which allowed us to reduce the dimensionality of the data to $d = 100$.  We adopt the experimental procedure used in \cite{karanam2015sparse} and randomly split the dataset into training and test sets. The training set is used to learn the XQDA metric, which is then applied to project the test data. For a given train-test split, we randomly select $L = 10$ frames per subject for the gallery and probe sets. This procedure is repeated $10$ times for each train-test split and the train-test split is itself repeated $10$ times (for a total of $100$ experiments), and we report averaged results. We select $\lambda = 10, \zeta = \tau = 0.5$ using cross-validation. For comparison, we report the performance of ISR using LOMO+XQDA features on the ILIDS-VID dataset even though \cite{lisanti2015person} did not consider this dataset nor this feature space.

The experimental results are shown in Table \ref{table:ilidsvid}. The experimental setup for the SR based RSM, ISR, and SRID is identical. The remaining methods in Table \ref{table:ilidsvid} use $L > 10$. Since the amount of information about the probe increases with increasing $L$, we argue that Table \ref{table:ilidsvid} is not biased towards SR based methods and provides a fair comparison of competing methods. The results show that RSM outperforms all competing algorithms, including the SR based ISR and SRID methods as well as algorithms based on nearest neighbors (NN) \cite{wu2016deep,zhao2013unsupervised,farenzena2010person}, RankSVM \cite{chapelle2010efficient}, etc. Despite the serious challenges presented by the ILIDS-VID dataset, RSM is able to achieve $96.6 \%$ accuracy at rank 1, which is $38.6 \%$ higher (in absolute terms) than the best reported result in the literature and $11.5 \%$ higher than ISR. This shows that there is significant discriminative information in the multi-shot re-id problem and considerable performance gains can be achieved by algorithms which take advantage of this. Moreover, the results in Table \ref{table:ilidsvid} show that the LOMO+XQDA feature space is well-suited for difficult re-id problems, since ISR is able to outperform all of the algorithms reported in the literature (with the exception of the proposed RSM) by simply adopting this feature space.

\subsection{Performance on PRID 2011}
The PRID 2011 dataset contains $200$ subjects with sequences from two camera views, with $5-675$ frames per subject per view. Unlike ILIDS-VID, PRID 2011 does not contain occlusions and thus makes re-id significantly \textit{easier}. As in the ILIDS-VID experiment, we use the LOMO feature space and learn a $d = 100$ XQDA embedding after splitting the dataset into training and test sets. We adopt the setup in \cite{karanam2015sparse,wu2016deep,you2016top}. For each train-test split, we select $L = 10$ frames per subject for the gallery and probe. This procedure is repeated $10$ times for each train-test split and the train-test split is itself repeated $10$ times (for a total of $100$ experiments), and we report averaged results. We select $\lambda = 10, \zeta = \tau = 0.5$ using cross-validation. As with ILIDS-VID, we report the performance of ISR using LOMO+XQDA features, despite the fact that \cite{lisanti2015person} did not consider this dataset nor this feature space. 

The experimental setup for the SR based RSM, ISR, and SRID is identical. The remaining methods use $L > 10$, which, as argued in Section \ref{sec:ILIDSVID}, does not prevent us from conducting a fair comparison between the competing methods. The experimental results are shown in Table \ref{table:ilidsvid}. The results show that RSM achieves $99.4 \%$ rank $1$ accuracy on this dataset, which is a $29.4 \%$ improvement (in absolute terms) over the best published result. Interestingly, ISR achieves $99.1\%$ rank $1$ accuracy on this dataset, despite the fact that ISR \textit{does not} model joint sparsity of $\lbrace \bm{\mati{X}} \rbrace_{i=1}^L$. This suggests that the SR re-id framework paired with the LOMO+XQDA feature space is robust enough to tackle the PRID 2011 dataset. 

\subsection{Performance on CAVIAR4ReID}
To show the performance of RSM independently of the LOMO+QXDA feature space and to better compare RSM with ISR \cite{lisanti2015person}, we evaluate the performance of RSM on the CAVIAR4ReID dataset using the WHOS \footnote{Made available by the authors of \cite{lisanti2015person} at \url{http://www.micc.unifi.it/lisanti/source-code/whos/}.} feature space \cite{lisanti2015person}. This dataset contains $72$ subjects, with sequences collected from one or two camera views. Each sequence contains $10$ frames, with the frames exhibiting significant resolution, lighting, and occlusion variation. In this instance, $d = 2960$ and $N = 360$. 

We follow the experimental setup in \cite{lisanti2015person} and randomly split the dataset into gallery and probe sets, with each set containing $L=5$ frames per subject. We repeat this process $50$ times and report averaged results. We select $\lambda = 1e-4, \zeta = \tau = 0.5$ using cross-validation. The experimental results are reported in Table \ref{table:caviar} \footnote{We do not report results for SRID because this dataset was not considered in \cite{karanam2015sparse} and the implementation is not publicly available.}. The experimental conditions for all of the methods reported are identical. The results show that RSM consistently outperforms ISR for all ranks, with the largest performance gap at rank $1$ where RSM outperforms ISR by $5.7 \%$ (absolute). Both RSM and ISR outperform approaches reported in the literature by a large margin. 

\vspace{-1em}
\subsection{Performance on VIPeR}
Finally, we report results for the VIPeR dataset, which is a single-shot re-id dataset containing $300$ subjects recorded from two views. To compare with ISR, we employ the WHOS feature space. Following \cite{lisanti2015person}, we randomly select $L = 1$ frame per subject for the gallery and probe, repeating the procedure $10$ times. Table \ref{table:ilids119} shows the rank $1$ accuracy results. The results show that RSM is competitive with state-of-the-art single-shot algorithms, even though RSM was primarily designed to tackle multi-shot problems.

\subsection{Why RSM?}
\label{sec:why}
\begin{figure}
\centering
\includegraphics[scale=0.25]{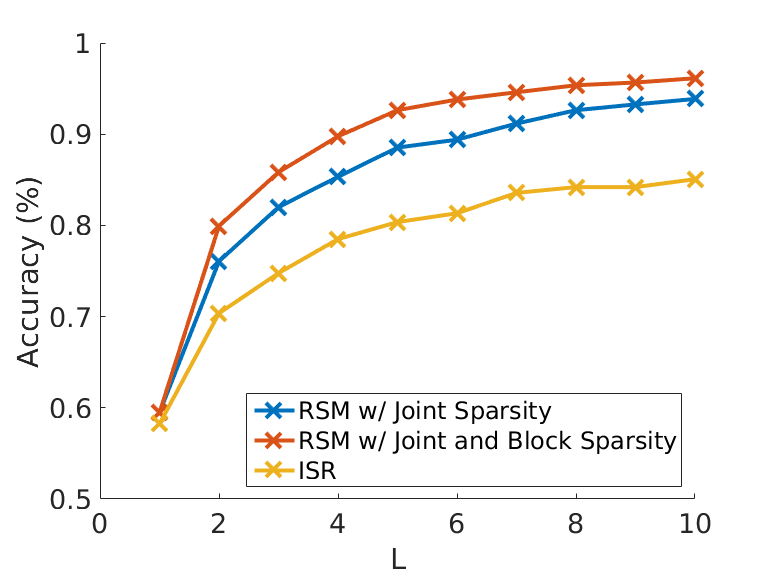}
\caption{ILIDS-VID rank 1 accuracy as a function of $L$.}
\label{fig:rank1 vs L}
\end{figure}

We now turn to the question of why RSM outperforms competing algorithms on the re-id problem. To motivate our arguments, we conduct a re-id experiment on the ILIDS-VID dataset using the LOMO+QXDA feature space and test the rank $1$ accuracy of RSM and ISR as a function of the number $L$ of gallery and probe frames per subject. To assess the influence of the various components of the RSM model on re-id performance, we show results for RSM with the joint and block sparsity components (i.e. the model detailed in Section \ref{sec:BSR}) as well as with the joint sparsity component but not the block sparsity component. The results are shown in Fig. \ref{fig:rank1 vs L}. First, we observe that even at $L=1$, RSM outperforms ISR by $1.2\%$ (in absolute terms). This provides evidence for the claim made in Section \ref{sec:intepretation} that the Type II inference procedure employed by RSM is superior to the Type I inference performed by ISR. Second, we see that there is an appreciable improvement in performance between RSM with joint sparsity and RSM with both joint and block sparsity. This suggests that each component of the RSM model provides a tangible improvement in re-id accuracy. Finally, the results show that the rank $1$ accuracy of RSM increases with increasing $L$, consistently outperforming ISR. In fact, the performance gap between RSM with joint and block sparsity and ISR \textit{increases} from $L=1$ to $L=6$, confirming the claim that RSM is \textit{better} able to harness the increasing amount of information about the probe subject as $L$ increases.

\begin{table}
\tiny
\begin{center}
\begin{tabular}{|ccc|}
\hline
Classifier & Feature space & Rank 1\\ \hline \hline 
RSM & WHOS \cite{lisanti2015person}  & $26.4$\\
ISR \cite{lisanti2015person} & WHOS \cite{lisanti2015person} & \bm{$27$} \\
NN & SDALF \cite{farenzena2010person} & $19.9$ \\
Bayes & CPS \cite{cheng2011custom}   & $21.8$ \\
RPLM \cite{hirzer2012relaxed} & HSV + LAB + LBP & \bm{$27$} \\
EIML \cite{hirzer2012person} & HSV + LAB + LBP   & $22$ \\
eSCD \cite{zhao2013unsupervised} & LAB + SIFT  & $26.7$ \\
\hline
\end{tabular}
\end{center}
\caption{Rank 1 accuracy ($\%$) on VIPeR dataset.}
\label{table:ilids119}
\end{table}



\vspace{-2em}

\section{Conclusion}
In this work, we presented the RSM re-id algorithm, which, to the best of our knowledge, is the first Bayesian SR based re-id algorithm. The RSM is capable of modeling all of the relevant aspects of the multi-shot re-id problem, including the block structure of the gallery $\bm{A}$, joint sparsity of the probe frames $\lbrace \bm{\mati{X}} \rbrace_{i=1}^L$, and sparse observation noise introduced by occlusions, viewpoint variations, and illuminated changes. In addition we introduced an efficient VB inference procedure and ranking scheme. We showed that RSM achieves state-of-the-art CMC results on every multi-shot re-id dataset considered and is competitive with existing approaches on the single-shot problem.

\vspace{-2em}

\bibliographystyle{IEEEtran}
\bibliography{IEEEabrv,egbib}
\vspace*{-3\baselineskip}
\begin{IEEEbiography}[{\includegraphics[width=1in,height=1.25in,clip,keepaspectratio]{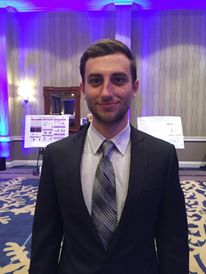}}]{Igor Fedorov}
(S'15) receieved the B.S and M.S. degrees in Electrical Engineering from the University of Illinois at Urbana-Champaign in 2012 and 2014, respectively. He is currently pursuing a Ph.D. degree in Electrical Engineering at the University of California, San-Diego. His research interests include sparse signal recovery, machine learning, and signal processing.
\end{IEEEbiography}
\vspace*{-3\baselineskip}
\begin{IEEEbiography}[{\includegraphics[width=1in,height=1.25in,clip,keepaspectratio]{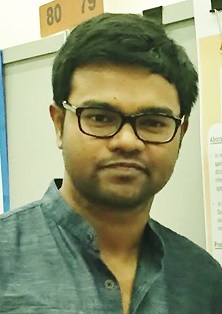}}]%
{Ritwik Giri}
(S'13) received B.E. degree in Electronics and Telecommunication Engineering from Jadavpur University, India in 2011. He also received M.S. and Ph.D. degrees in electrical engineering from the University of California, San Diego, La Jolla, CA, USA, in 2014 and 2016 respectively. He is currently working as a DSP Research Engineer at Starkey Hearing Technologies, Eden Prairie, MN, USA.
\end{IEEEbiography}
\vspace*{-3\baselineskip}
\begin{IEEEbiography}[{\includegraphics[width=1in,height=1.25in,clip,keepaspectratio]{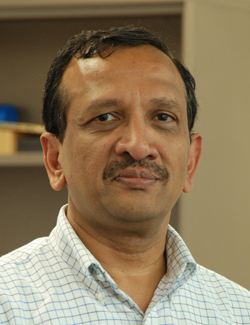}}]%
{Bhaskar D. Rao} (S'80-M'83-SM'91-F'00) received the B.Tech. degree in electronics and electrical communication engineering from the Indian Institute of Technology, Kharagpur, India, and the M.S. and Ph.D. degrees in electrical engineering from the University of Southern California, Los Angeles, Los Angeles, CA, USA, in 1979, 1981, and 1983, respectively. Since 1983, he has been with the University of California San Diego, La Jolla, CA, USA, where he is currently a Professor with the Department of Electrical and Computer Engineering and holds the Ericsson Endowed Chair in wireless access networks. He was the Director of the Center for Wireless Communications from 2008 to 2011.
\end{IEEEbiography}
\vspace*{-3\baselineskip}
\begin{IEEEbiography}[{\includegraphics[width=1in,height=1.25in,clip,keepaspectratio]{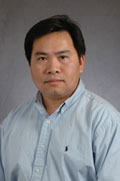}}]%
{Truong Q. Nguyen} (F’05) received the B.S., M.S.,
and Ph.D. degrees from California Institute of
Technology, Pasadena, CA, USA, in 1985, 1986,
and 1989, respectively.
He is a Professor with the Electrical and
Computer Engineering Department, University
of California at San Diego, San Diego, CA,
USA. He has co-authored a textbook entitled
Wavelets and Filter Banks (Wellesley-Cambridge,
1997; with Prof. G. Strang), and authored several
MATLAB-based toolboxes in image compression,
electrocardiogram compression, and filter bank design. He has also authored
over 400 publications. 
\end{IEEEbiography}

\end{document}